\documentclass[letterpaper]{article}
\usepackage[preprint]{aaai2027}
\usepackage[hyphens]{url}
\usepackage{graphicx}
\usepackage{amsmath}
\usepackage{natbib}
\usepackage{caption}
\usepackage{multirow}
\usepackage{booktabs}
\title{EpaCache: Error-Propagation-Aware Caching for Accelerating Diffusion-Based Visual Generation}
\author {
    Yuhan Liu,
    Zongwei Hong,
    Jinglun Li,
    Linze Li\textsuperscript{\(\ddagger\)},
    Shen Zhang,
    Yao Tang\textsuperscript{\(\dagger\)}
}
\affiliations {
    JIIOV Technology\\
    tedleo26@gmail.com,\\
    \{zongwei.hong, jinglun.li, shen.zhang, linze.li, yao.tang\}@jiiov.com
}

\begin{document}

\maketitle
\begingroup
\renewcommand{\thefootnote}{\fnsymbol{footnote}}
\footnotetext[2]{Corresponding author.}
\footnotetext[3]{Project leader.}
\endgroup

\begin{abstract}
Diffusion-based visual generative models deliver strong image and video synthesis quality but incur high inference costs because sequential samplers repeatedly evaluate large networks. Caching-based methods reduce inference latency by reusing intermediate computations across adjacent timesteps. However, existing cache controllers rely primarily on local temporal variation and overlook the trajectory-level consequences of cache reuse. We introduce Error-Propagation-Aware Cache (EpaCache), a training-free caching policy that adaptively allocates the reuse budget on timesteps with lower downstream impact. 
Experiments on image and video synthesis models demonstrate that EpaCache consistently improves the latency--fidelity trade-off over existing caching methods. On FLUX.1-dev, EpaCache outperforms the prior state-of-the-art caching method in both latency and fidelity, reducing inference time from $11.7$ s to $11.3$ s while improving PSNR from $21.4$ to $22.8$. On HunyuanVideo, EpaCache achieves a $2.63\times$ speedup over uncached inference and improves SSIM from $0.891$ to $0.905$ over the prior state-of-the-art method at matched latency.
\end{abstract}

\section{Introduction}
Recent visual generative models have achieved strong performance in image and video synthesis~\cite{opensora,opensora2,kong2024hunyuanvideo,flux2024,yang2024cogvideox,hong2022cogvideo,rombach2022high}. As model backbones evolve from U-Net architectures~\cite{ronneberger2015u} to Diffusion Transformer (DiT)~\cite{peebles2023scalable}, generation quality continues to improve, but inference becomes increasingly expensive because each sample is produced through a sequential multi-step sampling process. This repeated model evaluation is especially costly for high-resolution image and video generation, where latency, memory use, and computation directly limit practical deployment.

Existing acceleration techniques reduce this cost from different perspectives. Step reduction and distillation methods can shorten the sampling trajectory~\cite{salimans2022progressive,meng2023distillation,sauer2024adversarial}, while post-training quantization reduces the cost of individual model evaluations~\cite{chen2025q,liu2021post}. However, these methods often require additional training, extra fine-tuning data, or model-specific engineering.  Caching-based acceleration instead provides a complementary training-free alternative by reusing model outputs or intermediate computations across adjacent timesteps, thereby exploiting temporal redundancy along the sampling trajectory without modifying the original model weights. Existing caching methods~\cite{selvaraju2024forafastforwardcachingdiffusion,zhao2025realtimevideogenerationpyramid,zou_accelerating_2025,liu_timestep_2025,TaylorSeer2025,chung_seacache_2026} typically employ static thresholds, fixed skip intervals, or finely handcrafted rules to determine when cached computations should be reused, with their decisions driven primarily by temporal changes between adjacent sampling steps. 
\begin{figure}[t]
\centering
\includegraphics[width=\columnwidth
]{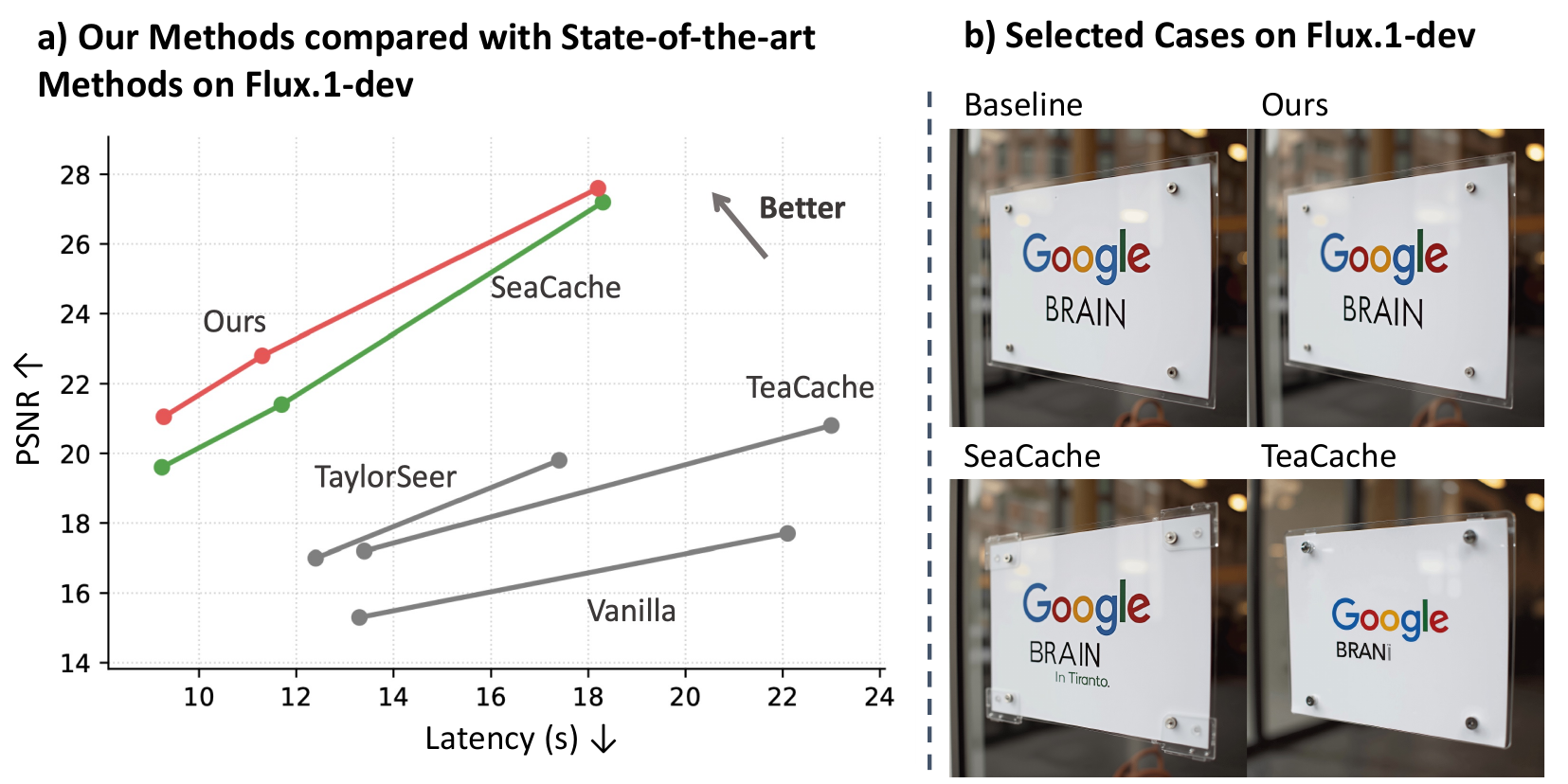}
\caption{(a) PSNR–latency trade-off compared with Vanilla, TaylorSeer, TeaCache, and SeaCache, where the upper-left direction indicates better performance. (b) A representative case on FLUX.1-dev.}
\label{fig:Introduction-Performance-Only}
\end{figure}
\begin{figure}[t]
\centering
\includegraphics[width=\columnwidth
]{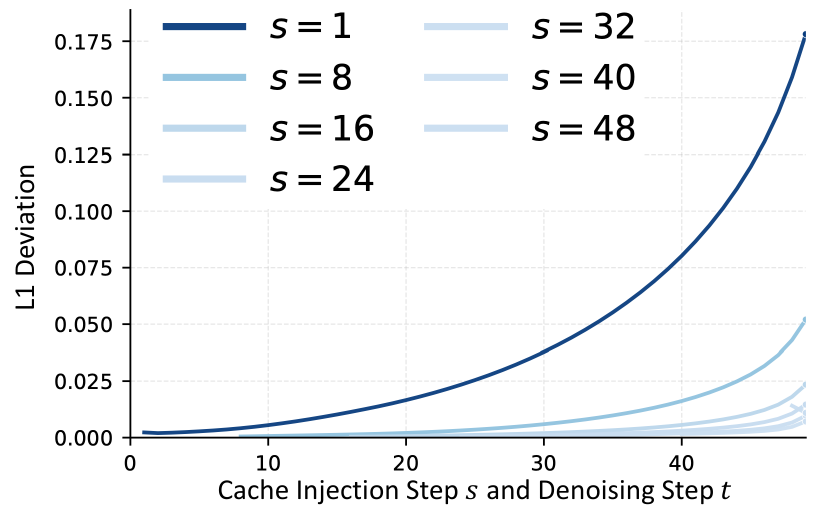}
\caption{L1 deviation trajectories caused by cache-induced approximation error at selected denoising steps $s$. Darker curves indicate larger terminal deviations.}
\label{fig:Introduction}
\end{figure}

However, such local reuse criteria do not explicitly account for how a cache-induced approximation propagates through the remaining sampling trajectory and ultimately affects generation quality. As observed in Fig.~\ref{fig:Introduction}, cache errors injected at different timesteps exhibit substantially different propagation effects. An error introduced early in the trajectory can be carried through many subsequent updates and may cause a large final deviation, whereas a cache injected later near the end has fewer opportunities to propagate. Moreover, at some timesteps, error caused by caching propagates more rapidly and can lead to larger final deviations than reuse at their neighboring timesteps. Therefore, the cost of cache reuse is not determined only by the local variation between adjacent steps; it also depends on how strongly an cache injected at that timestep affects the downstream trajectory. This observation raises a simple question: \textit{can we design a caching policy that explicitly accounts for timestep-dependent error propagation and allocates the caching budget accordingly?}

Motivated by this observation, we propose \textbf{EpaCache}, a training-free, error-propagation-aware caching policy. EpaCache calibrates the downstream impact of cache reuse at each cacheable timestep by injecting an isolated reuse perturbation and measuring the resulting final-state deviation. Given a user-specified global mean tolerance, it then redistributes the tolerance across timesteps: high-impact timesteps receive smaller thresholds and are refreshed more conservatively, whereas low-impact timesteps receive larger thresholds and permit more aggressive cache reuse. 

Our main contributions are threefold.
\begin{itemize}
    \item \textbf{Observation:} We identify and analyze error propagation in cache-based acceleration, showing that cache-induced approximation errors not only propagate through subsequent denoising steps but also vary substantially in their downstream impact depending on when they are introduced. 
    \item \textbf{Method:} We propose an error-propagation-aware caching policy, termed EpaCache. EpaCache adaptively assigns smaller reuse tolerances to high-impact timesteps and larger tolerances to low-impact timesteps, enabling efficient, high-fidelity diffusion model acceleration.
    \item \textbf{Performance:} Extensive experiments on multiple image and video generative models show that EpaCache reduces final deviation and improves reconstruction fidelity across different acceleration budgets, while maintaining a strong latency-quality trade-off over prior caching baselines.
\end{itemize}

\section{Related Work}
\subsection{Other Diffusion Model Acceleration Methods}
Diffusion and flow-matching models generate images and videos through iterative sampling trajectories that repeatedly evaluate a U-Net or diffusion transformer~\cite{ho2020denoisingdiffusionprobabilisticmodels,rombach2022high,peebles2023scalable,esser2024scaling}. Acceleration methods outside caching reduce either the number of model evaluations or the cost of each evaluation. Solver-based methods redesign the sampling trajectory while retaining the pretrained denoiser. DDIM permits deterministic sampling over a subset of timesteps, DPM-Solver uses diffusion-specific high-order updates, and UniPC adopts a unified predictor--corrector framework for few-step sampling~\cite{song2020denoising,lu2022dpm,zhao2023unipc}. Knowledge distillation instead transfers the behavior of a multi-step teacher to a student with a shorter trajectory. Progressive distillation repeatedly halves the required sampling steps, while guided and adversarial distillation improve the fidelity of few-step or one-step generation~\cite{salimans2022progressive,meng2023distillation,sauer2024adversarial}.

Complementary approaches reduce the cost of each denoising step. Quantization lowers the numerical precision of model weights and activations, while post-training quantization adapts quantization ranges to the timestep-dependent activation distributions of diffusion models without full retraining~\cite{li2023q,shang2023post,chen2025q,liu2021post}. Efficient attention improves memory access and parallelism for exact attention, whereas compressed or sparse attention reduces the number of token interactions in diffusion transformers~\cite{dao2022flashattention,dao2024flashattention,yuan2024ditfastattn,yang2026sparse,zhang2025sla}. Token-reduction methods further merge, prune, or dynamically select redundant spatial and temporal tokens so that subsequent transformer blocks process shorter sequences~\cite{zhang_training-free_2025,bolya2023token}. These methods provide substantial speedups, but may require additional training or optimization, alter the sampler or model computation graph, or rely on architecture- and hardware-specific implementations.

\subsection{Cache-Based Diffusion Model Acceleration}
Cache-based acceleration exploits temporal redundancy in iterative sampling by reusing model outputs or intermediate features across nearby timesteps. DeepCache first identified that high-level U-Net features evolve slowly along the diffusion trajectory and reused them across adjacent steps while updating inexpensive low-level features~\cite{ma2024deepcache}.

As diffusion transformers became the dominant backbone for visual generation, subsequent methods primarily exploited redundancy within DiT computations. FORA reuses attention and MLP outputs, while $\Delta$-DiT applies stage-dependent caching to front and rear transformer blocks~\cite{selvaraju2024forafastforwardcachingdiffusion,chen2024deltadittrainingfreeaccelerationmethod}. ToCa performs token-wise scoring to select which tokens should be cached or refreshed, while profiling-based methods preserve computation for heterogeneous blocks or visually dynamic foreground regions~\cite{zou_accelerating_2025,ma_model_2025}. For video DiTs, PAB broadcasts attention outputs at module-dependent intervals, FasterCache additionally reuses information across classifier-free guidance branches, and AdaCache adapts its schedule to video motion~\cite{zhao2025realtimevideogenerationpyramid,lv_fastercache_2025,kahatapitiya_adaptive_2024}.

Other adaptive methods improve either cache scheduling or the approximation of skipped features. DiCache uses shallow-layer online probes both to determine refresh timing and to combine multiple cached trajectories~\cite{bu2025dicache}, TaylorSeer forecasts future features through Taylor expansion and NaviCache tracks the evolving relationship between input and output variations at test time~\cite{TaylorSeer2025,lv2026navicache}.

Rather than evaluating the expensive denoiser output at every step, timestep-aware proxy-based caching methods estimate its temporal variation using lightweight timestep-dependent signals. TeaCache establishes the timestep-modulated input-proxy paradigm by modulating the noisy input with the timestep embedding and using the resulting input difference as a proxy for output change~\cite{liu_timestep_2025}. MagCache extends proxy-based cache control by modeling accumulated reuse error with the magnitude ratios of successive residual outputs~\cite{ma_magcache_2025}. SeaCache instead constructs a spectrally aligned input representation that suppresses noise-dominated components and measures redundancy in a content-relevant feature space~\cite{chung_seacache_2026}. BWCache is a structurally distinct instance of the same general principle: rather than constructing a model-level proxy from inputs or residual statistics, it directly compares block features at adjacent timesteps and triggers reuse independently for blocks whose feature differences fall below a threshold~\cite{cui_bwcache_2026}. Thus, these methods differ in proxy representation and granularity, but all use a low-cost temporal signal to determine whether current computation can be replaced by cached information.

\section{Methods}
\subsection{Preliminary}
\paragraph{\textbf{Rectified Flow and Flow Matching.}}
Flow matching~\citep{lipman2023flow,esser2024scaling} learns a continuous-time velocity field that transports samples between a simple prior and the data distribution. Let $\mathbf{x}_0$ denote a data sample and $\mathbf{x}_1 \sim \mathcal{N}(\mathbf{0}, \mathbf{I})$ denote a noise sample. A prescribed interpolant defines an intermediate state
\begin{equation}
\mathbf{x}_t = (1 - \rho(t))\,\mathbf{x}_0 + \rho(t)\,\mathbf{x}_1,
\end{equation}
where $t \in [0,1]$, $\rho(0)=0$, and $\rho(1)=1$. The target velocity along this path is
\begin{equation}
\mathbf{v}^{\ast}(\mathbf{x}_t,t)
=
\frac{\mathrm{d}\mathbf{x}_t}{\mathrm{d}t}
=
\dot{\rho}(t)\,(\mathbf{x}_1-\mathbf{x}_0).
\end{equation}
The model is trained to predict this velocity:
\begin{equation}
\mathcal{L}_{\mathrm{FM}} =
\mathrm{E}_{\mathbf{x}_0,\mathbf{x}_1,t}
\left[
\left\|
\mathbf{v}_{\theta}(\mathbf{x}_t,t) -
\mathbf{v}^{\ast}(\mathbf{x}_t,t)
\right\|_2^2
\right].
\end{equation}
At inference time, generation starts from the noise endpoint $\mathbf{x}_1$ and numerically integrates the learned velocity field backward from $t=1$ to $t=0$ to reach the data endpoint. Modern visual generators expose a sequence of adjacent sampling steps whose model evaluations are highly correlated, making them natural targets for caching.

Although diffusion and flow matching use different parameterizations, both produce an ordered sampling trajectory. For ease of interpretation, we report the discrete timestep index along the diffusion trajectory in this work.

\subsection{EpaCache: Error-Propagation-Aware Cache}
\label{sub:epacache_revised}

Rather than applying a static threshold uniformly across all timesteps, EpaCache allocates the caching budget to steps that have lower downstream impact. EpaCache decomposes cache control into a global-level timestep-specific reuse tolerance and an online local-variation signal. Given a user-specified global mean threshold $\bar{\delta}$, we derive a timestep-specific threshold $\delta_t$ based on the calibrated downstream impact of each timestep. High impact timesteps receive smaller thresholds and are refreshed more conservatively, whereas less impactful timesteps receive larger thresholds, enabling more aggressive cache reuse.

\begin{figure}[!ht]
\centering
\includegraphics[width=\columnwidth]{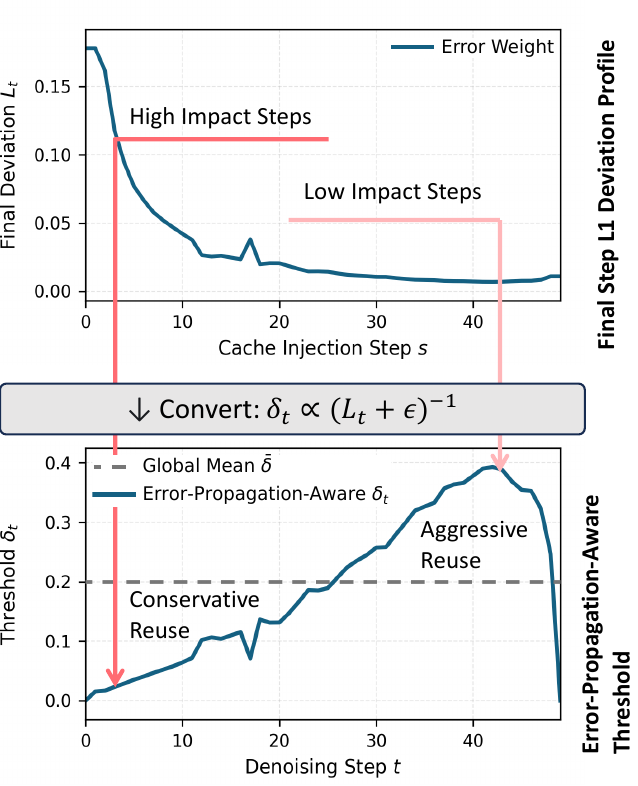}
\caption{Illustration of EpaCache's error-propagation-aware threshold allocation. By getting the downstream impact of each cacheable timestep over the calibration prompts, the inverse-normalized impact profile is scaled by the user-specified mean threshold $\bar{\delta}$ to obtain timestep-specific reuse thresholds. Therefore, high-impact timesteps receive smaller thresholds, whereas low-impact timesteps receive larger thresholds.}
\label{fig:method-global-level-threshold}
\end{figure}

\paragraph{\textbf{Error-Propagation-Aware Cache Budget Allocation.}}
We first estimate how much a cache-induced approximation error at each timestep affects the final sampling result and treat it as a timestep-aware caching impact. Let \(T\) denote the total number of sampling steps, and define the set of cacheable timesteps as $\mathcal{T}$ , namely, all sampling steps except the first and the last. The first step is excluded because no preceding residual is available for reuse, while the last step is fully computed to prevent severe quality degradation.

For each cacheable timestep \(t\in\mathcal{T}\), we replace the full model computation at \(t\) by reusing the residual from the preceding timestep \(t-1\), while retaining full computation at all subsequent timesteps. We define \(L_t\) as the average final-state relative L1 deviation $\ell_{1,\mathrm{rel}}$ between the perturbed and reference trajectories over $N$ prompts. $N$ is set to $20$ by default and was selected from T2V-CompBench~\cite{sun2024t2v}.

\begin{equation}
\label{eq:rel_l1}
\ell_{1,\mathrm{rel}}(A, B)
=
\frac{\lVert A-B\rVert_1}
{\lVert B\rVert_1+\xi}.
\end{equation}
\begin{equation}
\label{eq:definition-Lt}
L_t = \frac{1}{N}\sum_{n=1}^{N}
\ell_{1,\mathrm{rel}}(\widetilde{\mathbf{z}}_{\mathrm{final},t}^n, \mathbf{z}_{\mathrm{final}}^n)
\end{equation}

$\widetilde{\mathbf{z}}_{\mathrm{final},t}^n$ and $\mathbf{z}_{\mathrm{final}}^n$ represent the final latents of the perturbed trajectories where the cache is injected at the timestep $t$ and the reference trajectories, respectively. \(\xi\) is a small constant introduced for numerical stability. 

Repeating this procedure over all cacheable timesteps gives a final-deviation profile $\{L_t\}$, where a larger $L_t$ indicates that a local cache approximation error introduced at timestep $t$ has a stronger downstream impact and should therefore be handled more conservatively.

Given the calibrated final-deviation profile, we construct a timestep-specific threshold schedule that redistributes the user-specified global mean threshold across timesteps. Let $\bar{\delta}$ denote the global mean threshold, we assign smaller reuse tolerance to timesteps with larger final deviations and larger reuse tolerance to less impactful timesteps, while preserving the same mean reuse threshold.

We first convert the final deviation $L_t$ at each cacheable timestep into a cache allocation weight:

\begin{equation}
\label{eq:alpha_t-allocation}
\alpha_t=
\frac{(L_t+\xi)^{-1}}
{\frac{1}{|\mathcal{T}|}\sum_{s\in\mathcal{T}}(L_s+\xi)^{-1}},
\end{equation}
where $\xi$ is a small numerical stabilizer to prevent division by zero. The allocation weight $\alpha_t$ is inversely related to the final deviation, so timesteps with larger downstream impact receive smaller reuse tolerances.

We then obtain the timestep-specific threshold by scaling the user-specified average threshold:
\begin{equation}
\delta_t=\bar{\delta}\alpha_t.
\end{equation}

By construction, the weights have unit mean, and therefore the resulting thresholds satisfy
\begin{equation}
\frac{1}{|\mathcal{T}|}\sum_{t\in \mathcal{T}}\delta_t=\bar{\delta}.
\end{equation}

The reciprocal transformation in Eq.~\ref{eq:alpha_t-allocation} amplifies differences among low-impact timesteps, enabling finer discrimination in the low-deviation regime, which is common to late-stage sampling.

We further characterize the computational overhead of the calibration procedure and introduce an optional optimization strategy that substantially reduces the calibration cost. Detailed analysis and implementation details are provided in Appendix.

\begin{figure}[!ht]
\centering
\includegraphics[width=\columnwidth]{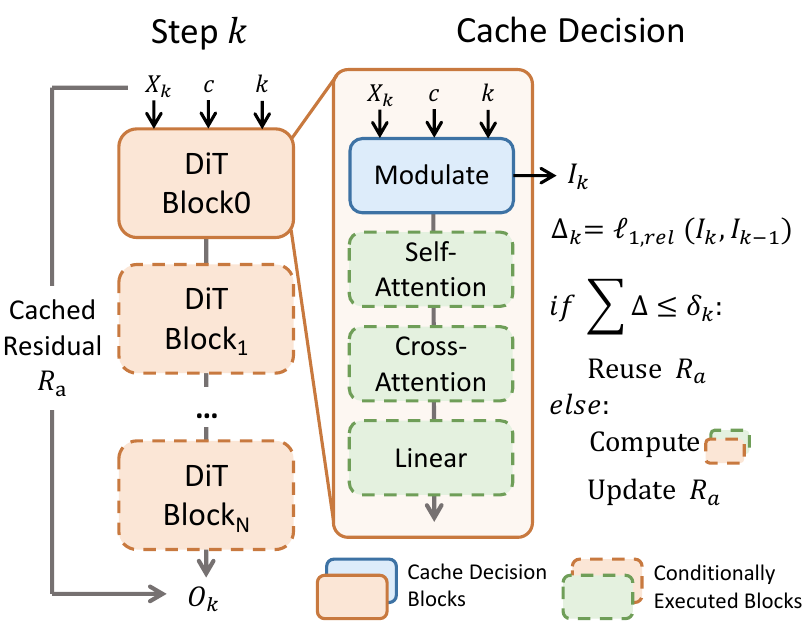}
\caption{Overview of the EpaCache caching proxy and the online cache decision. The adopted timestep-modulated proxy estimates the local variation \(\Delta_k\). The accumulated variation determines whether the cached residual $R_{a}$ is reused or the full DiT computation is conducted.}
\label{fig:method-local-variation}
\end{figure}

\paragraph{Local Variation Proxy and Residual Reuse.}
\label{sub:method-teacache}
To obtain a lightweight estimate of local model variation, we adopt the timestep-modulated input representation established in prior proxy-based caching work~\cite{liu_timestep_2025}. Let \(X_t\) denote the noisy input to the diffusion transformer $F_\theta$ at timestep \(t\), and let \(O_t=F_\theta(X_t,t,c)\) denote its output under condition \(c\). The proxy is the timestep-conditioned input representation \(I_t=modulate(X_t,t)\), obtained by modulating \(X_t\) with the timestep embedding. The temporal variation at a timestep is measured by the relative \(\ell_1\) distance
\begin{equation}
\label{eq:local-variation-rel-l1}
\Delta_t
=
\ell_{1,\mathrm{rel}}(I_t, I_{t-1})
=
\frac{\lVert I_t-I_{t-1}\rVert_1}
{\lVert I_{t-1}\rVert_1+\xi}.
\end{equation}
The temporal variation \(\Delta_t\) provides an inexpensive estimate of local change.

When the transformer is evaluated at a refresh iteration $a$, we cache its residual  
\begin{equation}
\label{eq:cached-residual}
R_{a}=O_{a}-X_{a}.
\end{equation}  
If the cache controller chooses to reuse the cached computation at a later iteration \(k>a\), the transformer output is approximated as  
\begin{equation}
\label{eq:residual-reuse}
\widetilde{O}_k=X_k+R_{a}.
\end{equation}  

The local variation proxy thus provides a lightweight estimate of how much the model state changes between consecutive diffusion iterations.
\paragraph{Online Cache Decision.}
With a timestep-specific error-propagation-aware threshold and an lightweight proxy to measure the local variation, during the diffusion iterations, EpaCache reuses the cached residual if
\begin{equation}
\label{eq:accum-th-epacache}
\sum_{j={a}+1}^{k}\Delta_j \leq \delta_k.
\end{equation}
Otherwise, the transformer is fully computed, and the residual $R_{a}$ is refreshed accordingly.

\section{Experiments}
\subsection{Experiment Settings}
\paragraph{\textbf{Model configurations.}} We evaluate three state-of-the-art visual generative models: FLUX.1-dev~\cite{labs2025flux1kontextflowmatching,flux2024} for text-to-image generation, Wan2.1-1.3B~\citep{wan2025} and HunyuanVideo~\cite{kong2024hunyuanvideo} for text-to-video generation. All experiments follow the default model configurations, except that HunyuanVideo generates 65 frames. HunyuanVideo is evaluated on NVIDIA H20 GPUs, FLUX.1-dev and Wan2.1 run on NVIDIA L20 GPUs.
\paragraph{\textbf{Dataset and Evaluation Protocol.}} For text-to-image generation, we evaluate 200 DrawBench prompts~\cite{saharia2022photorealistic} and generate images with resolution of $1024\times1024$. For text-to-video generation, we use 944 prompts from VBench~\cite{huang2023vbench}. For each model and prompt, the full-step output of the original sampler is used as the reference. We compute PSNR on RGB values, LPIPS~\cite{8578166}, and SSIM~\cite{1284395} between each cached output and its corresponding reference, and then average the metrics over all samples. The unit of PSNR is dB. The same initial random seed is shared across our method and all baselines. Inference latency is reported in seconds.
\paragraph{\textbf{Baseline configurations.}} To ensure a fair comparison, we mainly adjust the reuse tolerance to control the inference latency. For MagCache~\cite{ma_magcache_2025} and TaylorSeer~\cite{TaylorSeer2025}, we use the authors' default settings.
\subsection{Quantitative Comparison}

\begin{table}[t]
\centering
\small
\setlength{\tabcolsep}{1.8pt}
\begin{tabular}{@{}lrrrr@{}}
\toprule
Method & Latency (s) $\downarrow$ & PSNR $\uparrow$ & SSIM $\uparrow$ & LPIPS $\downarrow$ \\
\midrule
\multicolumn{5}{c}{\textit{FLUX.1-dev (image, $1024\times1024$)}} \\
\midrule
FLUX.1-dev~($T=50$) & 43.0 & -- & -- & -- \\
FLUX.1-dev~($T=25$) & 22.1 & 17.7 & 0.731 & 0.303 \\
FLUX.1-dev~($T=15$) & 13.3 & 15.3 & 0.648 & 0.420 \\
\midrule
TeaCache~($\delta=0.3$) & 23.0 & 20.8 & 0.807 & 0.199 \\
TaylorSeer~($\mathcal{S}=3$) & 17.4 & 19.8 & 0.780 & 0.222 \\
SeaCache~($\delta=0.3$) & 18.3 & 27.2 & 0.905 & \textbf{0.086} \\
EpaCache~($\bar{\delta}=0.2$) & 18.2 & \textbf{27.6} & \textbf{0.905} & 0.088 \\
\midrule
TeaCache~($\delta=0.6$) & 13.4 & 17.2 & 0.703 & 0.341 \\
TaylorSeer~($\mathcal{S}=5$) & 12.4 & 17.0 & 0.690 & 0.345 \\
SeaCache~($\delta=0.6$) & 11.7 & 21.4 & 0.802 & 0.211 \\
EpaCache~($\bar{\delta}=0.45$) & 11.3 & \textbf{22.8} & \textbf{0.812} & \textbf{0.207} \\
\bottomrule
\end{tabular}
\caption{Quantitative comparison on FLUX.1-dev under two efficiency regimes, corresponding to mean skip rates of approximately 50\% and 70\%. For TaylorSeer, we use the default approximation order $\mathcal{O}=2$.}
\label{tab:quantitative_flux}
\end{table}

\begin{figure}[t]
\centering
\includegraphics[width=\columnwidth]{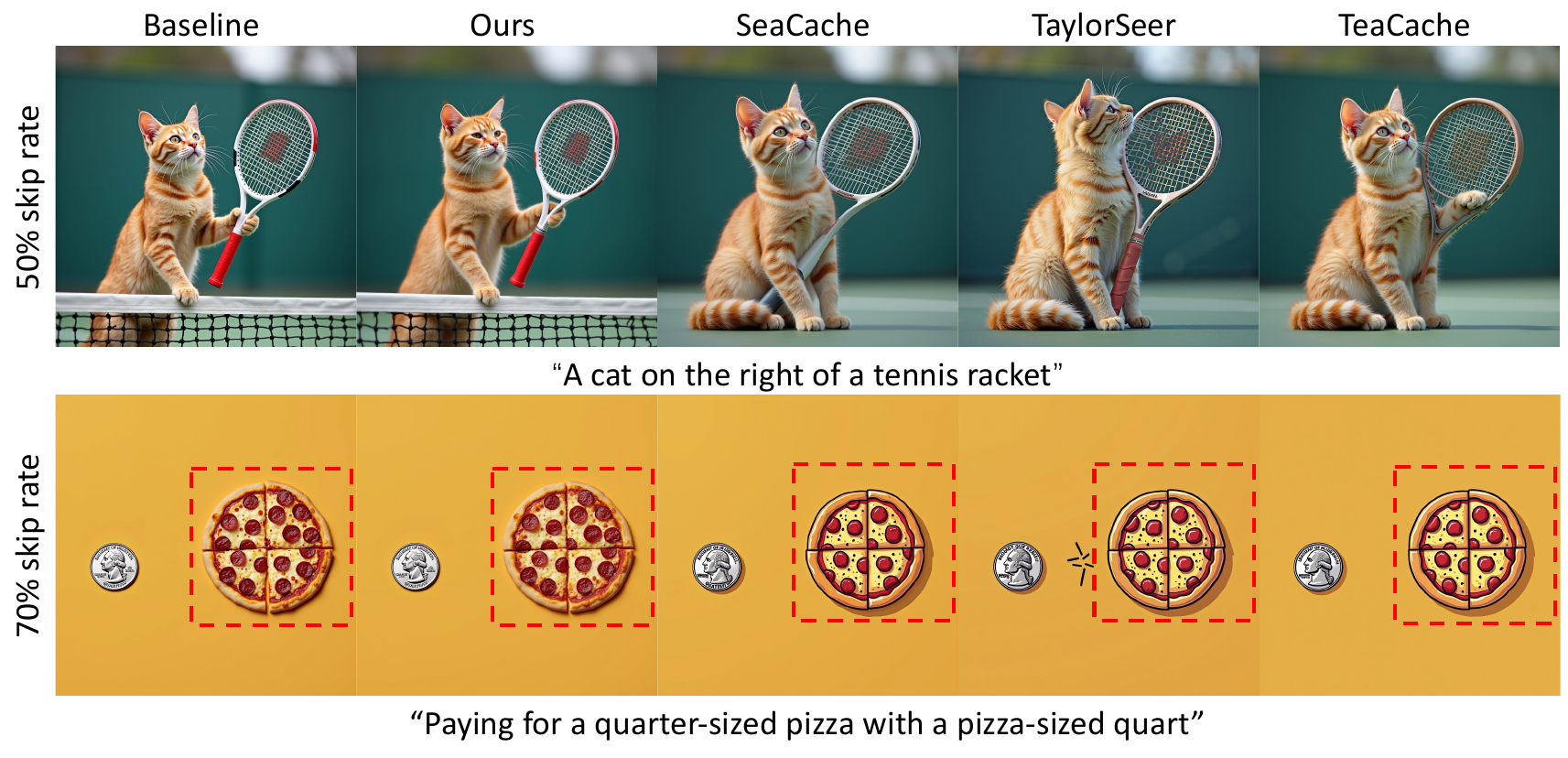}
\caption{Qualitative comparison of EpaCache and the baselines on FLUX.1-dev at skip rates of approximately 50\% and 70\%.}  
\label{fig:qualitative-flux}
\end{figure}

\begin{figure*}[!ht]
\centering
\includegraphics[width=\textwidth]{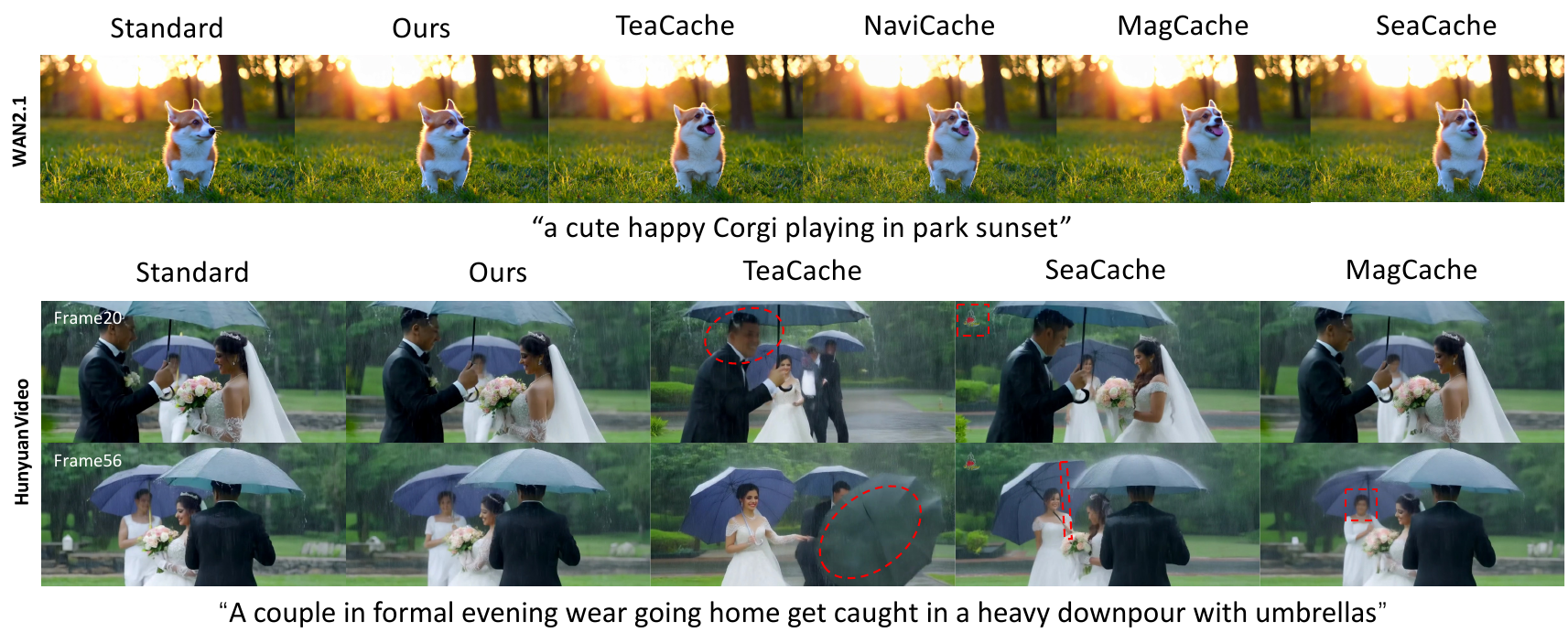}
\caption{Qualitative comparisons on Wan2.1 and HunyuanVideo. All methods are compared at matched frame indices; for HunyuanVideo, we show two representative frames within a video.}
\label{fig:viz_video}
\end{figure*}

\paragraph{\textbf{Text-to-image generation.}}
We compare EpaCache with existing caching methods on FLUX.1-dev~\citep{flux2024,labs2025flux1kontextflowmatching} under two efficiency regimes, with mean skip rates of approximately 50\% and 70\%, as shown in Tab.~\ref{tab:quantitative_flux}. Under the moderate regime, EpaCache substantially improves reconstruction fidelity over the faster TaylorSeer setting and achieves latency comparable to SeaCache, with the best PSNR, tied-best SSIM, and comparable LPIPS. This advantage persists under the more aggressive regime (approximately 70\% skip rate), where EpaCache achieves the lowest latency while maintaining the best overall fidelity.

\begin{table}[t]
\centering
\small
\setlength{\tabcolsep}{1.8pt}
\begin{tabular}{@{}lrrrr@{}}
\toprule
Method & Latency $\downarrow$ & PSNR $\uparrow$ & SSIM $\uparrow$ & LPIPS $\downarrow$ \\
\midrule
\multicolumn{5}{c}{\textit{Wan2.1-1.3B (81 frames, $832\times480$)}} \\
\midrule
Wan2.1-1.3B~($T=50$) & 409.55 & -- & -- & -- \\
Wan2.1-1.3B~($T=35$) & 289.05 & 16.74 & 0.6198 & 0.3105 \\
\midrule
TeaCache~($\delta=0.09$) & 214.65 & 22.36 & 0.7960 & 0.1388 \\
MagCache~($\delta=0.12$) & 205.09 & 28.04 & 0.9128 & 0.0510 \\
SeaCache~($\delta=0.20$) & 209.59 & 29.60 & 0.9222 & 0.0496 \\
NaviCache~($\tau=0.04I$) & 202.16 & 27.89 & 0.8924 & 0.0589	 \\
\midrule
\textbf{EpaCache}~($\bar{\delta}=0.06$) & 212.88 & \textbf{30.47} & \textbf{0.9346} & \textbf{0.0383} \\
\bottomrule
\end{tabular}
\caption{Quantitative comparison on Wan2.1-1.3B. \(\tau\) and \(I\) denote the error threshold and the identity matrix, respectively, which are hyperparameters in NaviCache~\cite{lv2026navicache}.}
\label{tab:quantitative_wan}
\end{table}

\begin{table*}[t] \centering \small \setlength{\tabcolsep}{2pt} 
\begin{tabular}{@{}l c rrrr rrr r@{}} 
\toprule
\multirow[c]{2}{*}{\textbf{Method}} &
\multirow[c]{2}{*}{\textbf{Venue}} &
\multicolumn{4}{c}{\textbf{Efficiency}} & \multicolumn{3}{c}{\textbf{Visual Retention}} & 
\multirow[c]{2}{*}{\textbf{VBench} $\uparrow$} \\ \cmidrule(lr){3-6} \cmidrule(lr){7-9} & & Latency (s) $\downarrow$ & Speedup $\uparrow$ & Memory (GB) $\downarrow$ & FLOPs (P) $\downarrow$ & PSNR $\uparrow$ & SSIM $\uparrow$ & LPIPS $\downarrow$ & \\ 
\midrule 

\multicolumn{10}{c}{\textit{HunyuanVideo (65 frames, $960\times544$)}} \\ 

\midrule HunyuanVideo~($T=50$) & -- & 753.84 & 1.00$\times$ & 45.60 & 24.904 & -- & -- & -- & 81.11 \\ 

HunyuanVideo~($T=25$) & -- & 392.44 & 1.92$\times$ & 45.60 & 13.031 & 20.16 & 0.7370 & 0.2759 & \underline{80.95} \\ 

\midrule TeaCache~($\delta=0.18$) & CVPR'25 & 287.40 & 2.62$\times$ & 46.20 & 9.580 & 21.59 & 0.7653 & 0.2378 & \textbf{81.14} \\ 

MagCache~($\delta=0.24$) & NeurIPS'25 & 285.11 & 2.64$\times$ & 45.80 & 9.505 & \textbf{29.44} & \underline{0.8965} & \textbf{0.0877} & 80.19 \\ 

SeaCache~($\delta=0.33$) & CVPR'26 & 290.69 & 2.59$\times$ & 46.59 & 9.688 & 28.71 & 0.8907 & 0.0971 & 80.65 \\ 

\midrule \textbf{EpaCache ($\bar{\delta}=0.25$)} & -- & 286.11 & 2.63$\times$ & 46.20 & 9.538 & \underline{29.25} & \textbf{0.9053} & \underline{0.0893} & 80.37 \\ \bottomrule
\end{tabular} 
\caption{ Efficiency, visual retention, and VBench comparison on HunyuanVideo. FLOPs are reported in peta floating-point operations (P). The best and second-best results for each metric are highlighted in \textbf{bold} and \underline{underline}, respectively. } 
\label{tab:main_hunyuan} \end{table*}

\paragraph{\textbf{Text-to-video generation.}}
Tabs.~\ref{tab:quantitative_wan} and~\ref{tab:main_hunyuan} compare EpaCache with existing methods on Wan2.1-1.3B and HunyuanVideo. On Wan2.1-1.3B, EpaCache achieves 30.47 PSNR at 212.88\,s, outperforming SeaCache in fidelity at similar latency and NaviCache in PSNR despite its slightly lower latency. On HunyuanVideo, EpaCache is faster than SeaCache (286.11\,s vs.\ 290.69\,s) while improving PSNR, SSIM, and LPIPS. Overall, EpaCache provides a favorable efficiency--fidelity trade-off across both video models.

\subsection{Qualitative Comparison}

\paragraph{\textbf{Text-to-image generation.}} In Fig.~\ref{fig:qualitative-flux}, we compare EpaCache with TeaCache~\cite{liu_timestep_2025}, TaylorSeer~\cite{TaylorSeer2025}, and SeaCache~\cite{chung_seacache_2026} at mean skip rates of approximately 50\% and 70\%. EpaCache preserves both the semantic content and overall perceptual quality of the original images, whereas the state-of-the-art methods frequently lose semantic details, or change the style of the original image. EpaCache faithfully preserves the content of the original images.

\paragraph{\textbf{Text-to-video generation.}}
We further conduct qualitative comparisons on the text-to-video models Wan2.1-1.3B~\cite{wan2025} and HunyuanVideo~\cite{kong2024hunyuanvideo}, as shown in Fig.~\ref{fig:viz_video}. For each prompt, we compare frames at the same temporal index across different methods. On Wan2.1-1.3B, the facial expression of the corgi generated by SeaCache also deviates noticeably from the baseline.  On HunyuanVideo, TeaCache tends to introduce excessive modifications to the semantic content of the generated frames. For example, in Frame 56, the position of the umbrella is noticeably inconsistent with the baseline output. Meanwhile, SeaCache introduces an undesirable watermark-like artifact in the upper-left corner.

\begin{figure*}[t]
\centering
\includegraphics[width=\textwidth]{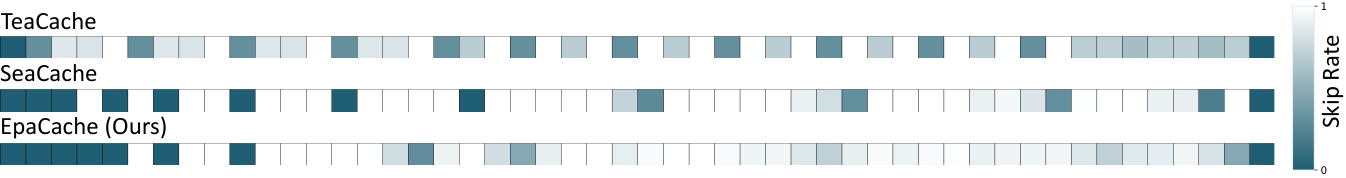}
\caption{Refresh pattern across timesteps on FLUX.1-dev on DrawBench. Each strip proceeds from the first to the last of the 50 denoising steps. Darker colors indicate lower skip probabilities and more frequent recomputation.}
\label{fig:anblation-skip-viz}
\end{figure*}

\subsection{Additional Analysis}

\begin{figure}[t]
\centering
\includegraphics[width=\columnwidth]{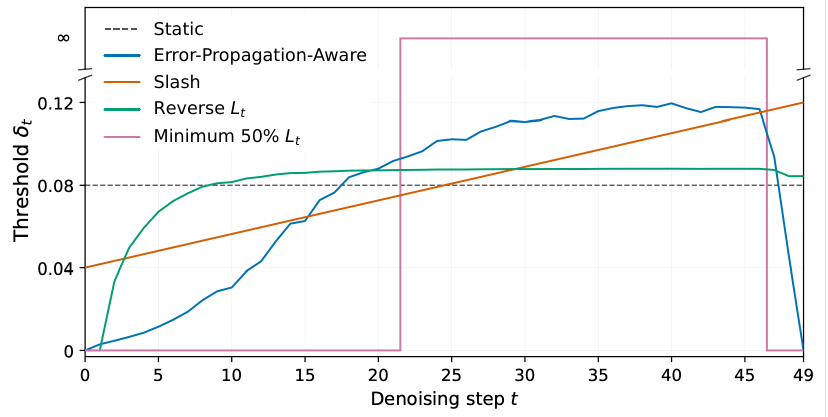}
\caption{Visualization of the per-step threshold-allocation schedules used in the ablation study. For Minimum 50\% $L_t$, the steps with the lowest 50\% of $L_t$ values are assigned an effectively infinite threshold (denoted by $\infty$), forcing cache reuse, while the remaining steps are assigned a threshold of zero.}
\label{fig:ablation-threshold-allocation}
\end{figure}

\begin{table}[!b]
\centering
\small
\setlength{\tabcolsep}{1.8pt}
\begin{tabular}{@{}lrrrr@{}}
\toprule
Line Shape & Latency $\downarrow$ & PSNR $\uparrow$ & SSIM $\uparrow$ & LPIPS $\downarrow$ \\
\midrule
Static($\delta=0.08$) & 209.61 & 21.04 & 0.7454 & 0.1798 \\
Minimum 50\% $L_t$ & 232.45 & 23.87 & 0.7412 & 0.1865 \\
Reverse $L_t$~($\bar{\delta}=0.08$) & 205.73 & 22.20 & 0.7799 & 0.1491 \\
Slash~($\bar{\delta}=0.08$) & 207.89 & 24.10 & 0.8310 & 0.1089 \\
\midrule
Ours~($\bar{\delta}=0.08$) & 215.49 & \textbf{28.33} & \textbf{0.9058} & \textbf{0.0577}	 \\
\bottomrule
\end{tabular}
\caption{Ablation study of threshold allocation on Wan2.1. We compare the proposed inverse-normalized Final-L1-weighted schedule with linear Slash schedules, a reverse $L_t$ schedule, and hard Minimum $50\%$ step-selection policies on 100 VBench prompts. For continuous schedules, the mean threshold is kept fixed when comparing curve shapes.}
\label{tab:quantitative_ablation_wan}
\end{table}

\paragraph{\textbf{Ablation study on threshold allocation.}}
We conduct a controlled ablation on Wan2.1 to isolate the contributions of the local-variation proxy and the temporal threshold-allocation rule in EpaCache. All variants are evaluated using the same 100-prompt VBench subset. As shown in Fig.~\ref{fig:ablation-threshold-allocation}, we compare our error-propagation-aware schedule, constructed from the final-L1 profile, against: (i) a static threshold, (ii) a linear slash schedule based on $\bar{\delta}$, (iii) a complementary schedule based on a revered version of $L_t$ and (iv) a hard selection rule that caches only the 50\% of steps with the lowest $L_t$. These comparisons reveal whether the local-variation proxy is necessary and how different threshold-curve shapes affect generation fidelity at their measured acceleration ratios.

As shown in Tab.~\ref{tab:quantitative_ablation_wan}, \textit{Minimum 50\% $L_t$} suffers a substantial performance degradation, demonstrating the necessity of a local proxy for identifying stale cached features. The inferior performance of \textit{Reverse $L_t$} further confirms the importance of assigning thresholds inversely proportional to feature importance. Although the \textit{Slash} strategy improves performance to some extent, it remains inferior to our Error-Propagation-Aware Threshold strategy.

\paragraph{\textbf{Sensitivity to the Number of Calibration Prompts.}}
To examine the effect of the calibration-set size, we vary the number of calibration prompts $N$ on FLUX.1-dev, as reported in Tab.~\ref{tab:calibrated_prompts_flux}. The results exhibit non-monotonic behavior when only a few prompts are used. In particular, $N=5$ substantially degrades reconstruction fidelity. Although $N=1$ happens to yield competitive results, an impact profile estimated from a single prompt can be highly prompt-dependent and should not be regarded as robust. Increasing $N$ to 10 or 20 produces more representative propagation estimates and a more stable latency--fidelity trade-off. We therefore use $N=20$ by default, which achieves the best overall fidelity with a modest calibration cost.

\begin{table}[t]
  \centering
  \small
  \setlength{\tabcolsep}{1.8pt}
  \begin{tabular}{@{}lrrrr@{}}
    \toprule
    $N$
    & Latency~(s)~\(\downarrow\)
    & PSNR~\(\uparrow\)
    & SSIM~\(\uparrow\)
    & LPIPS~\(\downarrow\) \\
    \midrule
    1~($\bar{\delta}=0.45$)  & 11.3 & 22.4 & 0.804 & 0.218 \\
    5~($\bar{\delta}=0.45$)  & 9.6 & 20.0 & 0.761 & 0.276 \\
    10~($\bar{\delta}=0.45$) & 11.1 & 22.3 & 0.804 & 0.216 \\
    20~($\bar{\delta}=0.45$) & 11.3 & \textbf{22.8} & \textbf{0.812} & \textbf{0.207} \\
    \bottomrule
  \end{tabular}
  \caption{Ablation study on the number of calibrated prompts $N$ using
  FLUX.1-dev.}
  \label{tab:calibrated_prompts_flux}
\end{table}

\paragraph{\textbf{Cache Decision Visualization.}}
Fig.~\ref{fig:anblation-skip-viz} compares the timestep-wise cache decision patterns of TeaCache~\cite{liu_timestep_2025}, SeaCache~\cite{chung_seacache_2026}, and our proposed EpaCache. TeaCache~($3.21\times$ speedup; $70.0\%$ skip rate) exhibits a largely regular and quasi-periodic refresh schedule, distributing its recomputation budget relatively uniformly across the denoising trajectory. SeaCache~($3.68\times$; $74.0\%$), in contrast, employs a Wiener-like spectral filter and front-loads recomputation during the early denoising steps, where global low-frequency structures are progressively established. As denoising proceeds, it increasingly reuses cached features, resulting in longer intervals with high cache rates. EpaCache~($3.82\times$; $74.7\%$) follows a different strategy by explicitly accounting for error propagation and assigning recomputation to timesteps whose approximation errors would have a greater downstream impact. Consequently, its refresh decisions are distributed across multiple influential regions rather than being restricted to a small set of fixed checkpoints.

\section{Conclusion}
In this paper, we introduced EpaCache, a training-free caching policy for accelerating diffusion-based visual generative models. By analyzing cache-induced errors, we showed that their downstream effects vary substantially across timesteps. EpaCache addresses this variation by calibrating timestep-wise propagation impacts with isolated cache-reuse perturbations and converting them into an adaptive threshold schedule, enabling conservative refreshes at sensitive timesteps and aggressive reuse elsewhere. Extensive experiments on FLUX.1-dev, Wan2.1, and HunyuanVideo demonstrated consistently improved latency--quality trade-offs over existing caching baselines under comparable realized computation budgets. These results highlight error-propagation-aware cache allocation as an effective strategy that can be integrated into existing cache controllers without model retraining or architectural modifications.

\bibliography{references}

@inproceedings{ronneberger2015u,
  title={U-net: Convolutional networks for biomedical image segmentation},
  author={Ronneberger, Olaf and Fischer, Philipp and Brox, Thomas},
  booktitle={International Conference on Medical image computing and computer-assisted intervention},
  pages={234--241},
  year={2015},
  organization={Springer}
}

@inproceedings{peebles2023scalable,
  title={Scalable diffusion models with transformers},
  author={Peebles, William and Xie, Saining},
  booktitle={Proceedings of the IEEE/CVF international conference on computer vision},
  pages={4195--4205},
  year={2023}
}

@inproceedings{ma2024deepcache,
  title={Deepcache: Accelerating diffusion models for free},
  author={Ma, Xinyin and Fang, Gongfan and Wang, Xinchao},
  booktitle={Proceedings of the IEEE/CVF conference on computer vision and pattern recognition},
  pages={15762--15772},
  year={2024}
}

@misc{liu_timestep_2025,
	title = {Timestep {Embedding} {Tells}: {It}'s {Time} to {Cache} for {Video} {Diffusion} {Model}},
	url = {http://arxiv.org/abs/2411.19108},
	doi = {10.48550/arXiv.2411.19108},
	author = {Liu, Feng and Zhang, Shiwei and Wang, Xiaofeng and Wei, Yujie and Qiu, Haonan and Zhao, Yuzhong and Zhang, Yingya and Ye, Qixiang and Wan, Fang},
	month = mar,
	year = {2025},
	note = {arXiv:2411.19108 [cs]},
}

@misc{selvaraju2024forafastforwardcachingdiffusion,
      title={FORA: Fast-Forward Caching in Diffusion Transformer Acceleration}, 
      author={Pratheba Selvaraju and Tianyu Ding and Tianyi Chen and Ilya Zharkov and Luming Liang},
      year={2024},
      eprint={2407.01425},
      archivePrefix={arXiv},
      primaryClass={cs.CV},
      url={https://arxiv.org/abs/2407.01425}, 
}

@misc{zhao2025realtimevideogenerationpyramid,
      title={Real-Time Video Generation with Pyramid Attention Broadcast}, 
      author={Xuanlei Zhao and Xiaolong Jin and Kai Wang and Yang You},
      year={2025},
      eprint={2408.12588},
      archivePrefix={arXiv},
      primaryClass={cs.CV},
      url={https://arxiv.org/abs/2408.12588}, 
}

@misc{zhang_training-free_2025,
	title = {Training-{Free} {Efficient} {Video} {Generation} via {Dynamic} {Token} {Carving}},
	url = {http://arxiv.org/abs/2505.16864},
	doi = {10.48550/arXiv.2505.16864},
	author = {Zhang, Yuechen and Xing, Jinbo and Xia, Bin and Liu, Shaoteng and Peng, Bohao and Tao, Xin and Wan, Pengfei and Lo, Eric and Jia, Jiaya},
	month = nov,
	year = {2025},
	note = {arXiv:2505.16864 [cs]},
}

@misc{zou_accelerating_2025,
	title = {Accelerating {Diffusion} {Transformers} with {Token}-wise {Feature} {Caching}},
	url = {http://arxiv.org/abs/2410.05317},
	doi = {10.48550/arXiv.2410.05317},
	author = {Zou, Chang and Liu, Xuyang and Liu, Ting and Huang, Siteng and Zhang, Linfeng},
	month = feb,
	year = {2025},
	note = {arXiv:2410.05317 [cs]},
}

@misc{lv_fastercache_2025,
	title = {{FasterCache}: {Training}-{Free} {Video} {Diffusion} {Model} {Acceleration} with {High} {Quality}},
	url = {http://arxiv.org/abs/2410.19355},
	doi = {10.48550/arXiv.2410.19355},
	author = {Lv, Zhengyao and Si, Chenyang and Song, Junhao and Yang, Zhenyu and Qiao, Yu and Liu, Ziwei and Wong, Kwan-Yee K.},
	month = mar,
	year = {2025},
	note = {arXiv:2410.19355 [cs]},
}

@misc{ma_magcache_2025,
	title = {{MagCache}: {Fast} {Video} {Generation} with {Magnitude}-{Aware} {Cache}},
	url = {http://arxiv.org/abs/2506.09045},
	doi = {10.48550/arXiv.2506.09045},
	author = {Ma, Zehong and Wei, Longhui and Wang, Feng and Zhang, Shiliang and Tian, Qi},
	month = nov,
	year = {2025},
	note = {arXiv:2506.09045 [cs]},
}

@misc{kahatapitiya_adaptive_2024,
	title = {Adaptive {Caching} for {Faster} {Video} {Generation} with {Diffusion} {Transformers}},
	url = {http://arxiv.org/abs/2411.02397},
	doi = {10.48550/arXiv.2411.02397},
	author = {Kahatapitiya, Kumara and Liu, Haozhe and He, Sen and Liu, Ding and Jia, Menglin and Zhang, Chenyang and Ryoo, Michael S. and Xie, Tian},
	month = nov,
	year = {2024},
	note = {arXiv:2411.02397 [cs]},
}

@misc{chen2024deltadittrainingfreeaccelerationmethod,
      title={$\Delta$-DiT: A Training-Free Acceleration Method Tailored for Diffusion Transformers}, 
      author={Pengtao Chen and Mingzhu Shen and Peng Ye and Jianjian Cao and Chongjun Tu and Christos-Savvas Bouganis and Yiren Zhao and Tao Chen},
      year={2024},
      eprint={2406.01125},
      archivePrefix={arXiv},
      primaryClass={cs.CV},
      url={https://arxiv.org/abs/2406.01125}, 
}

@misc{chung_seacache_2026,
	title = {{SeaCache}: {Spectral}-{Evolution}-{Aware} {Cache} for {Accelerating} {Diffusion} {Models}},
	url = {http://arxiv.org/abs/2602.18993},
	doi = {10.48550/arXiv.2602.18993},
	author = {Chung, Jiwoo and Hyun, Sangeek and Lee, MinKyu and Han, Byeongju and Cha, Geonho and Wee, Dongyoon and Hong, Youngjun and Heo, Jae-Pil},
	month = mar,
	year = {2026},
	note = {arXiv:2602.18993 [cs]},
}

@misc{cui_bwcache_2026,
	title = {{BWCache}: {Accelerating} {Video} {Diffusion} {Transformers} through {Block}-{Wise} {Caching}},
	url = {http://arxiv.org/abs/2509.13789},
	doi = {10.48550/arXiv.2509.13789},
	author = {Cui, Hanshuai and Tang, Zhiqing and Xu, Zhifei and Yao, Zhi and Zeng, Wenyi and Jia, Weijia},
	month = feb,
	year = {2026},
	note = {arXiv:2509.13789 [cs]},
}

@misc{ma_model_2025,
	title = {Model {Reveals} {What} to {Cache}: {Profiling}-{Based} {Feature} {Reuse} for {Video} {Diffusion} {Models}},
	url = {https://arxiv.org/abs/2504.03140},
	doi = {10.48550/ARXIV.2504.03140},
	author = {Ma, Xuran and Liu, Yexin and Liu, Yaofu and Wu, Xianfeng and Zheng, Mingzhe and Wang, Zihao and Lim, Ser-Nam and Yang, Harry},
	year = {2025},
}

@article{TaylorSeer2025,
  title={From Reusing to Forecasting: Accelerating Diffusion Models with TaylorSeers},
  author={Liu, Jiacheng and Zou, Chang and Lyu, Yuanhuiyi and Chen, Junjie and Zhang, Linfeng},
  journal={arXiv preprint arXiv:2503.06923},
  year={2025}
}

@article{bu2025dicache,
  title={Dicache: Let diffusion model determine its own cache},
  author={Bu, Jiazi and Ling, Pengyang and Zhou, Yujie and Wang, Yibin and Zang, Yuhang and Lin, Dahua and Wang, Jiaqi},
  journal={arXiv preprint arXiv:2508.17356},
  year={2025}
}

@article{lv2026navicache,
  title={NaviCache: Test-Time Self-Calibration Caching for Video Generation},
  author={Lv, Zheqi and Zhu, Zhibo and Wang, Jinke and Tian, Qi and Zhang, Shengyu and Chen, Zhengyu and Zang, Chengxi and Zhao, Zhou and Wu, Fei},
  journal={arXiv preprint arXiv:2606.26795},
  year={2026}
}

@misc{labs2025flux1kontextflowmatching,
      title={FLUX.1 Kontext: Flow Matching for In-Context Image Generation and Editing in Latent Space},
      author={Black Forest Labs and Stephen Batifol and Andreas Blattmann and Frederic Boesel and Saksham Consul and Cyril Diagne and Tim Dockhorn and Jack English and Zion English and Patrick Esser and Sumith Kulal and Kyle Lacey and Yam Levi and Cheng Li and Dominik Lorenz and Jonas Müller and Dustin Podell and Robin Rombach and Harry Saini and Axel Sauer and Luke Smith},
      year={2025},
      eprint={2506.15742},
      archivePrefix={arXiv},
      primaryClass={cs.GR},
      url={https://arxiv.org/abs/2506.15742},
}

@misc{flux2024,
    author={Black Forest Labs},
    title={FLUX},
    year={2024},
    howpublished={\url{https://github.com/black-forest-labs/flux}},
}

@article{yang2024cogvideox,
  title={CogVideoX: Text-to-Video Diffusion Models with An Expert Transformer},
  author={Yang, Zhuoyi and Teng, Jiayan and Zheng, Wendi and Ding, Ming and Huang, Shiyu and Xu, Jiazheng and Yang, Yuanming and Hong, Wenyi and Zhang, Xiaohan and Feng, Guanyu and others},
  journal={arXiv preprint arXiv:2408.06072},
  year={2024}
}

@article{hong2022cogvideo,
  title={CogVideo: Large-scale Pretraining for Text-to-Video Generation via Transformers},
  author={Hong, Wenyi and Ding, Ming and Zheng, Wendi and Liu, Xinghan and Tang, Jie},
  journal={arXiv preprint arXiv:2205.15868},
  year={2022}
}

@article{kong2024hunyuanvideo,
  title={Hunyuanvideo: A systematic framework for large video generative models},
  author={Kong, Weijie and Tian, Qi and Zhang, Zijian and Min, Rox and Dai, Zuozhuo and Zhou, Jin and Xiong, Jiangfeng and Li, Xin and Wu, Bo and Zhang, Jianwei and others},
  journal={arXiv preprint arXiv:2412.03603},
  year={2024}
}

@article{opensora,
  title={Open-sora: Democratizing efficient video production for all},
  author={Zheng, Zangwei and Peng, Xiangyu and Yang, Tianji and Shen, Chenhui and Li, Shenggui and Liu, Hongxin and Zhou, Yukun and Li, Tianyi and You, Yang},
  journal={arXiv preprint arXiv:2412.20404},
  year={2024}
}

@article{opensora2,
    title={Open-Sora 2.0: Training a Commercial-Level Video Generation Model in \$200k}, 
    author={Xiangyu Peng and Zangwei Zheng and Chenhui Shen and Tom Young and Xinying Guo and Binluo Wang and Hang Xu and Hongxin Liu and Mingyan Jiang and Wenjun Li and Yuhui Wang and Anbang Ye and Gang Ren and Qianran Ma and Wanying Liang and Xiang Lian and Xiwen Wu and Yuting Zhong and Zhuangyan Li and Chaoyu Gong and Guojun Lei and Leijun Cheng and Limin Zhang and Minghao Li and Ruijie Zhang and Silan Hu and Shijie Huang and Xiaokang Wang and Yuanheng Zhao and Yuqi Wang and Ziang Wei and Yang You},
    year={2025},
    journal={arXiv preprint arXiv:2503.09642},
}

@article{wan2025,
      title={Wan: Open and Advanced Large-Scale Video Generative Models}, 
      author={Team Wan and Ang Wang and Baole Ai and Bin Wen and Chaojie Mao and Chen-Wei Xie and Di Chen and Feiwu Yu and Haiming Zhao and Jianxiao Yang and Jianyuan Zeng and Jiayu Wang and Jingfeng Zhang and Jingren Zhou and Jinkai Wang and Jixuan Chen and Kai Zhu and Kang Zhao and Keyu Yan and Lianghua Huang and Mengyang Feng and Ningyi Zhang and Pandeng Li and Pingyu Wu and Ruihang Chu and Ruili Feng and Shiwei Zhang and Siyang Sun and Tao Fang and Tianxing Wang and Tianyi Gui and Tingyu Weng and Tong Shen and Wei Lin and Wei Wang and Wei Wang and Wenmeng Zhou and Wente Wang and Wenting Shen and Wenyuan Yu and Xianzhong Shi and Xiaoming Huang and Xin Xu and Yan Kou and Yangyu Lv and Yifei Li and Yijing Liu and Yiming Wang and Yingya Zhang and Yitong Huang and Yong Li and You Wu and Yu Liu and Yulin Pan and Yun Zheng and Yuntao Hong and Yupeng Shi and Yutong Feng and Zeyinzi Jiang and Zhen Han and Zhi-Fan Wu and Ziyu Liu},
      journal = {arXiv preprint arXiv:2503.20314},
      year={2025}
}

@inproceedings{rombach2022high,
  title={High-resolution image synthesis with latent diffusion models},
  author={Rombach, Robin and Blattmann, Andreas and Lorenz, Dominik and Esser, Patrick and Ommer, Bj{\"o}rn},
  booktitle={Proceedings of the IEEE/CVF conference on computer vision and pattern recognition},
  pages={10684--10695},
  year={2022}
}

@misc{ho2020denoisingdiffusionprobabilisticmodels,
      title={Denoising Diffusion Probabilistic Models}, 
      author={Jonathan Ho and Ajay Jain and Pieter Abbeel},
      year={2020},
      eprint={2006.11239},
      archivePrefix={arXiv},
      primaryClass={cs.LG},
      url={https://arxiv.org/abs/2006.11239}, 
}

@article{song2020denoising,
  title={Denoising diffusion implicit models},
  author={Song, Jiaming and Meng, Chenlin and Ermon, Stefano},
  journal={arXiv preprint arXiv:2010.02502},
  year={2020}
}

@article{lu2022dpm,
  title={Dpm-solver: A fast ode solver for diffusion probabilistic model sampling in around 10 steps},
  author={Lu, Cheng and Zhou, Yuhao and Bao, Fan and Chen, Jianfei and Li, Chongxuan and Zhu, Jun},
  journal={Advances in neural information processing systems},
  volume={35},
  pages={5775--5787},
  year={2022}
}

@INPROCEEDINGS{8578166,
  author={Zhang, Richard and Isola, Phillip and Efros, Alexei A. and Shechtman, Eli and Wang, Oliver},
  booktitle={2018 IEEE/CVF Conference on Computer Vision and Pattern Recognition}, 
  title={The Unreasonable Effectiveness of Deep Features as a Perceptual Metric}, 
  year={2018},
  volume={},
  number={},
  pages={586-595},
  doi={10.1109/CVPR.2018.00068}
}

@ARTICLE{1284395,
  author={Zhou Wang and Bovik, A.C. and Sheikh, H.R. and Simoncelli, E.P.},
  journal={IEEE Transactions on Image Processing}, 
  title={Image quality assessment: from error visibility to structural similarity}, 
  year={2004},
  volume={13},
  number={4},
  pages={600-612},
  doi={10.1109/TIP.2003.819861}
}

@inproceedings{
lipman2023flow,
title={Flow Matching for Generative Modeling},
author={Yaron Lipman and Ricky T. Q. Chen and Heli Ben-Hamu and Maximilian Nickel and Matthew Le},
booktitle={The Eleventh International Conference on Learning Representations },
year={2023},
url={https://openreview.net/forum?id=PqvMRDCJT9t}
}

@article{zhao2023unipc,
  title={Unipc: A unified predictor-corrector framework for fast sampling of diffusion models},
  author={Zhao, Wenliang and Bai, Lujia and Rao, Yongming and Zhou, Jie and Lu, Jiwen},
  journal={Advances in Neural Information Processing Systems},
  volume={36},
  pages={49842--49869},
  year={2023}
}

@InProceedings{huang2023vbench,
    title={{VBench}: Comprehensive Benchmark Suite for Video Generative Models},
    author={Huang, Ziqi and He, Yinan and Yu, Jiashuo and Zhang, Fan and Si, Chenyang and Jiang, Yuming and Zhang, Yuanhan and Wu, Tianxing and Jin, Qingyang and Chanpaisit, Nattapol and Wang, Yaohui and Chen, Xinyuan and Wang, Limin and Lin, Dahua and Qiao, Yu and Liu, Ziwei},
    booktitle={Proceedings of the IEEE/CVF Conference on Computer Vision and Pattern Recognition},
    year={2024}
}

@article{saharia2022photorealistic,
  title={Photorealistic text-to-image diffusion models with deep language understanding},
  author={Saharia, Chitwan and Chan, William and Saxena, Saurabh and Li, Lala and Whang, Jay and Denton, Emily L and Ghasemipour, Kamyar and Gontijo Lopes, Raphael and Karagol Ayan, Burcu and Salimans, Tim and others},
  journal={Advances in neural information processing systems},
  volume={35},
  pages={36479--36494},
  year={2022}
}

@inproceedings{esser2024scaling,
  title={Scaling rectified flow transformers for high-resolution image synthesis},
  author={Esser, Patrick and Kulal, Sumith and Blattmann, Andreas and Entezari, Rahim and M{\"u}ller, Jonas and Saini, Harry and Levi, Yam and Lorenz, Dominik and Sauer, Axel and Boesel, Frederic and others},
  booktitle={Forty-first international conference on machine learning},
  year={2024}
}

@article{li2023q,
  title={Q-dm: An efficient low-bit quantized diffusion model},
  author={Li, Yanjing and Xu, Sheng and Cao, Xianbin and Sun, Xiao and Zhang, Baochang},
  journal={Advances in neural information processing systems},
  volume={36},
  pages={76680--76691},
  year={2023}
}

@inproceedings{shang2023post,
  title={Post-training quantization on diffusion models},
  author={Shang, Yuzhang and Yuan, Zhihang and Xie, Bin and Wu, Bingzhe and Yan, Yan},
  booktitle={Proceedings of the IEEE/CVF conference on computer vision and pattern recognition},
  pages={1972--1981},
  year={2023}
}

@article{dao2022flashattention,
  title={Flashattention: Fast and memory-efficient exact attention with io-awareness},
  author={Dao, Tri and Fu, Dan and Ermon, Stefano and Rudra, Atri and R{\'e}, Christopher},
  journal={Advances in neural information processing systems},
  volume={35},
  pages={16344--16359},
  year={2022}
}

@inproceedings{dao2024flashattention,
  title={Flashattention-2: Faster attention with better parallelism and work partitioning},
  author={Dao, Tri},
  booktitle={International Conference on Learning Representations},
  volume={2024},
  pages={35549--35562},
  year={2024}
}

@article{yuan2024ditfastattn,
  title={Ditfastattn: Attention compression for diffusion transformer models},
  author={Yuan, Zhihang and Zhang, Hanling and Lu, Pu and Ning, Xuefei and Zhang, Linfeng and Zhao, Tianchen and Yan, Shengen and Dai, Guohao and Wang, Yu},
  journal={Advances in Neural Information Processing Systems},
  volume={37},
  pages={1196--1219},
  year={2024}
}

@article{yang2026sparse,
  title={Sparse videogen2: Accelerate video generation with sparse attention via semantic-aware permutation},
  author={Yang, Shuo and Xi, Haocheng and Zhao, Yilong and Li, Muyang and Zhang, Jintao and Cai, Han and Lin, Yujun and Li, Xiuyu and Xu, Chenfeng and Peng, Kelly and others},
  journal={Advances in Neural Information Processing Systems},
  volume={38},
  pages={96965--96991},
  year={2026}
}

@article{zhang2025sla,
  title={Sla: Beyond sparsity in diffusion transformers via fine-tunable sparse-linear attention},
  author={Zhang, Jintao and Wang, Haoxu and Jiang, Kai and Yang, Shuo and Zheng, Kaiwen and Xi, Haocheng and Wang, Ziteng and Zhu, Hongzhou and Zhao, Min and Stoica, Ion and others},
  journal={arXiv preprint arXiv:2509.24006},
  year={2025}
}

@article{salimans2022progressive,
  title={Progressive distillation for fast sampling of diffusion models},
  author={Salimans, Tim and Ho, Jonathan},
  journal={arXiv preprint arXiv:2202.00512},
  year={2022}
}

@inproceedings{meng2023distillation,
  title={On distillation of guided diffusion models},
  author={Meng, Chenlin and Rombach, Robin and Gao, Ruiqi and Kingma, Diederik and Ermon, Stefano and Ho, Jonathan and Salimans, Tim},
  booktitle={Proceedings of the IEEE/CVF conference on computer vision and pattern recognition},
  pages={14297--14306},
  year={2023}
}

@inproceedings{sauer2024adversarial,
  title={Adversarial diffusion distillation},
  author={Sauer, Axel and Lorenz, Dominik and Blattmann, Andreas and Rombach, Robin},
  booktitle={European Conference on Computer Vision},
  pages={87--103},
  year={2024},
  organization={Springer}
}

@inproceedings{chen2025q,
  title={Q-dit: Accurate post-training quantization for diffusion transformers},
  author={Chen, Lei and Meng, Yuan and Tang, Chen and Ma, Xinzhu and Jiang, Jingyan and Wang, Xin and Wang, Zhi and Zhu, Wenwu},
  booktitle={Proceedings of the Computer Vision and Pattern Recognition Conference},
  pages={28306--28315},
  year={2025}
}

@article{liu2021post,
  title={Post-training quantization for vision transformer},
  author={Liu, Zhenhua and Wang, Yunhe and Han, Kai and Zhang, Wei and Ma, Siwei and Gao, Wen},
  journal={Advances in Neural Information Processing Systems},
  volume={34},
  pages={28092--28103},
  year={2021}
}

@inproceedings{bolya2023token,
  title={Token merging for fast stable diffusion},
  author={Bolya, Daniel and Hoffman, Judy},
  booktitle={Proceedings of the IEEE/CVF conference on computer vision and pattern recognition},
  pages={4599--4603},
  year={2023}
}

@article{sun2024t2v,
  title={T2V-CompBench: A Comprehensive Benchmark for Compositional Text-to-video Generation},
  author={Sun, Kaiyue and Huang, Kaiyi and Liu, Xian and Wu, Yue and Xu, Zihan and Li, Zhenguo and Liu, Xihui},
  journal={arXiv preprint arXiv:2407.14505},
  year={2024}
}

\appendix
\setcounter{secnumdepth}{1}
\twocolumn[{
\centering
{\LARGE\bfseries Appendix\par}
}]

This supplementary material provides additional results for the image model
FLUX.1-dev~\cite{flux2024} and the text-to-video models Wan2.1~\cite{wan2025}, and HunyuanVideo~\cite{kong2024hunyuanvideo}.

\section{Overhead of Error-Propagation Profiling}

The error-propagation-aware profile is constructed only once for each model and could be adapted to different threshold configurations for different acceleration scenarios. 

\begin{figure}[t]
\centering
\includegraphics[width=\columnwidth]{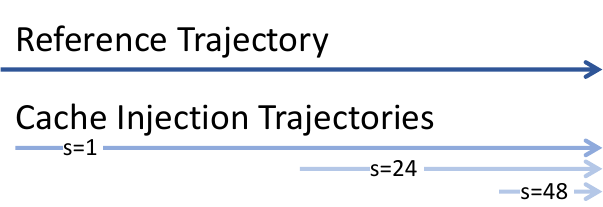}
\caption{Visualization of the error-propagation-aware calibration steps. For a cache injected at a timestep $s$, the remaining $T-s$ denoising steps are executed to evaluate the effect of cache reuse.}
\label{fig:viz-calibration-steps}
\end{figure}

As shown in Fig~\ref{fig:viz-calibration-steps}, we first perform a fullstep inference to obtain a reference trajectory. To evaluate the effect of caching at timestep $s$, we reuse the cached residual from timestep $s-1$ at timestep $s$ and resume sampling from the corresponding recorded state. Therefore, each profiling run executes only the remaining $T-s$ denoising steps instead of restarting the entire trajectory, where $T$ denotes the total number of sampling steps.

We exclude the first and last denoising steps from the set of cacheable timesteps. The first step has no preceding residual available for reuse, while the last step is always recomputed to preserve the final generation quality. Accordingly, the cacheable timestep set $\mathcal{C}$ is defined as
\begin{equation}
\label{eq:cacheable-timestep-set}
    \mathcal{C} = \{2, \ldots, T-1\},
    \qquad
    |\mathcal{C}| = T-2.
\end{equation}
Let $c_{\mathrm{step}}$ denote the computational cost of one denoising step,
which is constant across timesteps. The cost of one standard
$T$-step inference is
\begin{equation}
\label{eq:standard-inference-cost}
    C_{\mathrm{std}} = T c_{\mathrm{step}}.
\end{equation}
where $T=50$ is typically used in FLUX.1-dev~\cite{flux2024,labs2025flux1kontextflowmatching}, Wan2.1~\cite{wan2025} and HunyuanVideo~\cite{kong2024hunyuanvideo}. 

For a single calibration prompt, profiling includes the initial baseline
generation and all cache-injection runs. Its cost is therefore
\begin{equation}
\label{eq:per-prompt-profiling-cost}
\begin{aligned}
    C_{\mathrm{per-prompt}}
    &= \left[T+\sum_{s\in\mathcal{C}}(T-s)\right]c_{\mathrm{step}} \\
    &= \left[T+\frac{(T-2)(T-1)}{2}\right]c_{\mathrm{step}}.
\end{aligned}
\end{equation}
Given $N$ calibration prompts, the total offline calibration cost
$C_{\mathrm{per-model}}$ is
\begin{equation}
\label{eq:total-calibration-cost}
\begin{aligned}
    C_{\mathrm{per-model}} &= N \times C_{\mathrm{per-prompt}},
\end{aligned}
\end{equation}
and the ratio to one standard inference $\rho_{\mathrm{cal}}$ are
\begin{equation}
\label{eq:total-standard-inference-ratio}
\begin{aligned}
    \rho_{\mathrm{cal}}
    &= \frac{C_{\mathrm{per-model}}}{C_{\mathrm{std}}}
     = N\left[1+\frac{(T-2)(T-1)}{2T}\right].
\end{aligned}
\end{equation}
For FLUX with $T=50$, this ratio becomes
\begin{equation}
\label{eq:flux-calibration-cost}
    \rho_{\mathrm{cal}}
    = N\left(1+\frac{48\times49}{2\times50}\right)
    = 24.52N.
\end{equation}

Thus, profiling one prompt costs the equivalent of $24.52$ standard inference
runs, while the default calibration set of $N=20$ prompts requires approximately
$490.4$ standard-inference equivalents in total. The profiling runs are
independent across both timesteps and prompts and can therefore be executed in
parallel, which could be substantially faster in multi-GPU or distributed settings.

\section{Optimization: Reducing Profiling Overhead with Sparse-Step Interpolation}
\begin{table*}[t]
\centering
\small
\setlength{\tabcolsep}{4pt}
\begin{tabular}{@{}lrrrrr@{}}
\toprule
Scheme
& \shortstack{Per-Prompt Cost\\($\times$) $\downarrow$}
& Latency (s) $\downarrow$
& PSNR $\uparrow$
& SSIM $\uparrow$
& LPIPS $\downarrow$ \\
\midrule
\multicolumn{6}{c}{\textit{FLUX.1-dev (image, $1024\times1024$)}} \\
\midrule
Standard inference & 1.00 & 43.0 & -- & -- & -- \\
\midrule
\multicolumn{6}{l}{$\bar{\delta}=0.20$} \\
Full calibration & 24.52 & 18.0 & 27.6 & 0.905 & 0.088 \\
Interpolation ($1/2$ steps) & 12.26 & 18.2 & 27.9 & 0.905 & 0.085 \\
Interpolation ($1/3$ steps) & 8.17 & 18.8 & 28.0 & 0.910 & 0.083 \\
Interpolation ($1/4$ steps) & 6.13 & 18.3 & 28.1 & 0.911 & 0.081 \\
Interpolation ($1/5$ steps) & 4.90 & 18.5 & 28.2 & 0.911 & 0.082 \\
Interpolation ($1/8$ steps) & 3.07 & 18.4 & 28.4 & 0.915 & 0.078 \\
Interpolation ($1/10$ steps) & 2.45 & 18.3 & 27.9 & 0.910 & 0.084 \\
\midrule
\multicolumn{6}{l}{$\bar{\delta}=0.45$} \\
Full calibration & 24.52 & 11.3 & 22.8 & 0.812 & 0.207 \\
Interpolation ($1/2$ steps) & 12.26 & 11.5 & 22.7 & 0.810 & 0.210 \\
Interpolation ($1/3$ steps) & 8.17 & 11.5 & 22.8 & 0.811 & 0.208 \\
Interpolation ($1/4$ steps) & 6.13 & 11.1 & 22.7 & 0.810 & 0.210 \\
Interpolation ($1/5$ steps) & 4.90 & 11.3 & 22.9 & 0.811 & 0.209 \\
Interpolation ($1/8$ steps) & 3.07 & 11.7 & 23.0 & 0.818 & 0.210 \\
Interpolation ($1/10$ steps) & 2.45 & 11.0 & 22.6 & 0.808 & 0.213 \\

\bottomrule
\end{tabular}
\caption{Effect of sparse-step calibration on EpaCache for FLUX.1-dev.
The reported cost is the per-prompt profiling cost relative to one standard
inference; the total cost for $N$ calibration prompts is $N$ times the
reported value.}
\label{tab:optimize-reduce-calinration-steps}
\end{table*}

Although error-propagation profiling is performed only once offline for each model, it requires timestep-wise cache-injection runs for multiple cacheable timesteps. This may limit the practical applicability of EpaCache.

To reduce this overhead, we explore sparse-step calibration, which explicitly calibrates only a subset of the cacheable timesteps and interpolates final deviation $L_t$ allocations for the remaining steps.

We evaluate multiple interpolation ratios, including $1/2$, $1/3$, $1/4$,
$1/5$, $1/8$, and $1/10$, where a ratio of $1/k$ means that one out of every
$k$ timesteps is explicitly calibrated and the remaining profile values are
interpolated. The results are summarized in
Table~\ref{tab:optimize-reduce-calinration-steps}.

Across the tested ratios and global thresholds, the interpolated
error-propagation profiles maintain stable latency and reconstruction quality.
These results demonstrate that interpolation preserves the profile's ability
to control model performance and the resulting latency--quality trade-off,
while reducing the required profiling overhead.

\section{Dimension-Level VBench Evaluation on HunyuanVideo}
As reported in Table~\ref{tab:main_hunyuan} of the main paper, EpaCache
achieves an overall VBench score of 80.37 on HunyuanVideo, compared with 81.11
and 80.95 for the original 50-step and 25-step models, respectively. We further
compare the aggregate Semantic and Quality Scores and their constituent VBench
dimensions in Tables~\ref{tab:vbench_semantic_dimensions}
and~\ref{tab:vbench_quality_dimensions}.

\begin{table*}[p]
\centering
\small
\setlength{\tabcolsep}{4pt}
\begin{tabular}{@{}lrrrrrr@{}}
\toprule
Semantic Dimension $\uparrow$
& \shortstack{Original\\(50 steps)}
& \shortstack{Original\\(25 steps)}
& EpaCache
& TeaCache
& MagCache
& SeaCache \\
\midrule
Object Class         & 79.27 & 79.35 & \textbf{79.11} & 77.69 & 75.95 & \underline{78.09} \\
Multiple Objects     & 72.79 & 66.08 & \underline{73.09} & 69.28 & 71.88 & \textbf{74.16} \\
Human Action         & 92.00 & 92.00 & \textbf{92.00} & \underline{90.00} & 88.00 & 88.00 \\
Color                & 90.96 & 93.92 & 90.70 & \textbf{91.81} & \underline{91.40} & 91.05 \\
Spatial Relationship & 70.64 & 67.14 & 67.09 & \textbf{75.88} & 64.99 & \underline{67.16} \\
Scene                & 30.05 & 33.06 & 28.46 & \underline{30.76} & \textbf{31.29} & 29.43 \\
Appearance Style     & 66.34 & 66.73 & 66.51 & \textbf{67.36} & \underline{66.94} & 66.58 \\
Temporal Style       & 63.19 & 62.99 & \underline{62.88} & \textbf{63.52} & 62.66 & 62.64 \\
Overall Consistency  & 70.85 & 70.96 & \underline{70.27} & \textbf{70.36} & 69.97 & 70.25 \\
\midrule
Semantic Score       & 70.68 & 70.25 & \underline{70.01} & \textbf{70.74} & 69.23 & 69.71 \\
\bottomrule
\end{tabular}
\caption{Dimension-level VBench Semantic Scores on the 540p, 65-frame
evaluation setting. Bold and underlined entries denote the best and
second-best results, respectively, among the four caching methods; the
original-model columns are excluded from this comparison.}
\label{tab:vbench_semantic_dimensions}

\begin{tabular}{@{}lrrrrrr@{}}
\toprule
Quality Dimension $\uparrow$
& \shortstack{Original\\(50 steps)}
& \shortstack{Original\\(25 steps)}
& EpaCache
& TeaCache
& MagCache
& SeaCache \\
\midrule
Subject Consistency    & 95.84 & 95.84 & \underline{95.88} & 95.80 & \textbf{95.89} & 95.84 \\
Background Consistency & 96.78 & 97.05 & \textbf{96.97} & 96.21 & \underline{96.90} & 96.78 \\
Temporal Flickering    & 98.79 & 99.06 & \textbf{98.89} & 98.81 & \underline{98.87} & \underline{98.87} \\
Motion Smoothness      & 98.63 & 98.94 & \underline{98.70} & 98.46 & \textbf{98.73} & \textbf{98.73} \\
Aesthetic Quality      & 61.45 & 60.77 & \underline{61.18} & \textbf{61.25} & 60.56 & 61.11 \\
Imaging Quality        & 65.61 & 62.78 & \textbf{64.02} & \underline{63.95} & 62.42 & 63.61 \\
Dynamic Degree         & 27.08 & 29.17 & 25.70 & \textbf{29.86} & 25.70 & \underline{27.08} \\
\midrule
Quality Score          & 83.72 & 83.63 & 83.28 & \textbf{83.74} & 82.93 & \underline{83.39} \\
\bottomrule
\end{tabular}
\caption{Dimension-level VBench Quality Scores on the 540p, 65-frame
evaluation setting. Bold and underlined entries denote the best and
second-best results, respectively, among the four caching methods; the
original-model columns are excluded from this comparison.}
\label{tab:vbench_quality_dimensions}

\begin{tabular}{@{}lrrrrrr@{}}
\toprule
Average Rank $\downarrow$
& \shortstack{Original\\(50 steps)}
& \shortstack{Original\\(25 steps)}
& EpaCache
& TeaCache
& MagCache
& SeaCache \\
\midrule
Semantic & -- & -- & \underline{2.56} & \textbf{1.78} & 2.94 & 2.72 \\
Quality  & -- & -- & \textbf{1.93} & 2.86 & 2.64 & \underline{2.57} \\
Overall  & -- & -- & \underline{2.28} & \textbf{2.25} & 2.81 & 2.66 \\
\bottomrule
\end{tabular}
\caption{Average ranks among the four caching methods across the constituent
VBench dimensions. Semantic and Quality average ranks are computed over 9 and
7 dimensions, respectively, and Overall averages all 16 dimensions. Aggregate
Semantic and Quality Scores are excluded to avoid double counting, and ties
receive their average ranks. Bold and underlined entries denote the best and
second-best average ranks, respectively. The original-model columns are retained
for reference, with ranks shown as ``--'' because they are excluded from the
ranking.}
\label{tab:vbench_average_ranks}
\end{table*}

EpaCache achieves a Semantic Score of 70.01, 0.67 points below the 50-step
original model and the second-highest score among the caching methods. It
obtains the highest Object Class score (79.11) among the caching methods,
matches the original model on Human Action (92.00), and exceeds the 50-step
original model on Multiple Objects and Appearance Style.

EpaCache obtains a Quality Score of 83.28, only 0.44 points below the 50-step
original model. Among the caching methods, it achieves the best Background
Consistency (96.97), Temporal Flickering (98.89), and Imaging Quality (64.02),
as well as the best average rank across all Quality dimensions (1.93) in
Table~\ref{tab:vbench_average_ranks}.

As summarized in Table~\ref{tab:vbench_average_ranks}, EpaCache ranks first
among the caching methods on the Quality dimensions with an average rank of
1.93. It ranks second overall with an average rank of 2.28, only 0.03 behind
TeaCache.

\begin{figure*}[t]
\centering
\includegraphics[width=\textwidth]{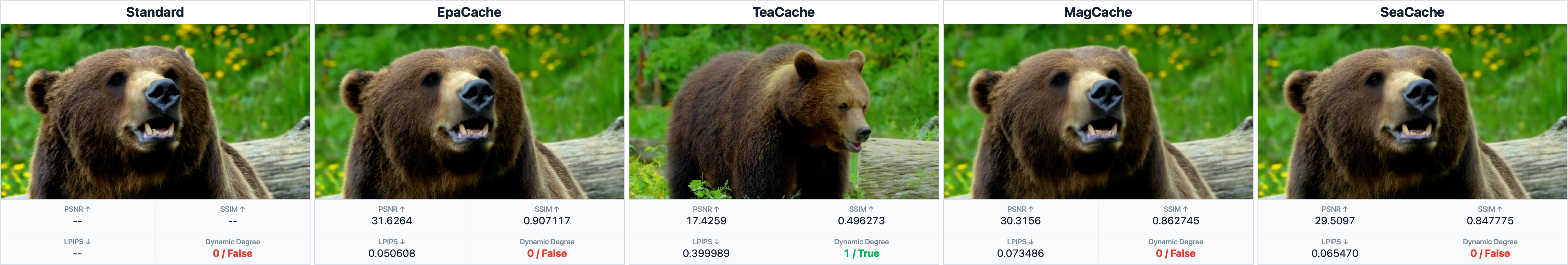}
\caption{Qualitative comparison of video generated by the
uncached reference, EpaCache, TeaCache, MagCache, and SeaCache with the prompt \textit{"a bear sniffing the air for scents of food"}. The prompt measures the VBench metric \textbf{Dynamic Degree}. The figure
also reports reference-fidelity metrics and the Dynamic Degree decision for
each caching method.}
\label{fig:dynamic-degree-qualitative}
\end{figure*}

\FloatBarrier
\section{Effectiveness of the Dynamic Degree Metric}

The VBench Dynamic Degree metric measures whether a generated video contains
sufficiently noticeable motion. However, cache-induced errors can manifest as
temporal flicker, semantic drift from the uncached output, or degradation of
perceptual quality. Because Dynamic Degree primarily captures the presence and
magnitude of motion, these failure modes may change the score without
consistently reflecting semantic preservation or visual fidelity.
Consequently, Dynamic Degree alone provides limited evidence for evaluating
training-free acceleration methods for visual generation and should be
considered alongside semantic, temporal-quality, and reference-fidelity
metrics. 

\subsubsection{Results on the Current Evaluation Set}

Table~\ref{tab:dynamic_degree_comparison} compares
TeaCache and EpaCache on the same $72$ videos used for
Dynamic Degree evaluation.

\begin{table*}[t]
    \centering
    \small
    \setlength{\tabcolsep}{5pt}
    \begin{tabular}{@{}lccc@{}}
        \toprule
        \textbf{Metric}
        &
        \textbf{TeaCache}
        &
        \textbf{EpaCache}
        &
        \textbf{Better Method}
        \\
        \midrule
        Dynamic Degree $\uparrow$
        & $43/72=0.5972$
        & $37/72=0.5139$
        & TeaCache
        \\
        Dynamic agreement with Standard $\uparrow$
        & $64/72=88.9\%$
        & $68/72=94.4\%$
        & EpaCache
        \\
        PSNR $\uparrow$
        & $21.31$
        & $29.49$
        & EpaCache
        \\
        SSIM $\uparrow$
        & $0.7127$
        & $0.9021$
        & EpaCache
        \\
        LPIPS $\downarrow$
        & $0.2357$
        & $0.0720$
        & EpaCache
        \\
        Subject Consistency $\uparrow$
        & $0.964116$
        & $0.964839$
        & EpaCache
        \\
        Motion Smoothness $\uparrow$
        & $0.993022$
        & $0.993739$
        & EpaCache
        \\
        \bottomrule
    \end{tabular}
    \caption{Comparison between Dynamic Degree and other quality metrics.}
    \label{tab:dynamic_degree_comparison}
\end{table*}

Although EpaCache has a lower Dynamic Degree score, it performs better in terms of PSNR, SSIM, LPIPS, Subject Consistency, and
Motion Smoothness. Most importantly, EpaCache has better allignment with the Standard generation, as measured by Dynamic Agreement, indicating that it better preserves the motion decisions of the uncached reference, instead of generating incorrct cotents.

These results demonstrate that a higher Dynamic Degree score may not
necessarily correspond to better video generation quality.

\subsubsection{Qualitative Visualization}

We additionally visualize representative outputs from EpaCache and TeaCache
in Fig.~\ref{fig:dynamic-degree-qualitative}. The visual comparison indicates
that EpaCache produces better overall perceptual quality than TeaCache in these
examples, with generated content and appearance that remain closer to the
uncached reference. This qualitative result further illustrates that a higher
Dynamic Degree score does not necessarily correspond to better visual quality.

\section{Supplementary Visualization and Case Study}
\label{app:visualization}

\begin{figure*}[t]
\centering
\includegraphics[width=\textwidth]{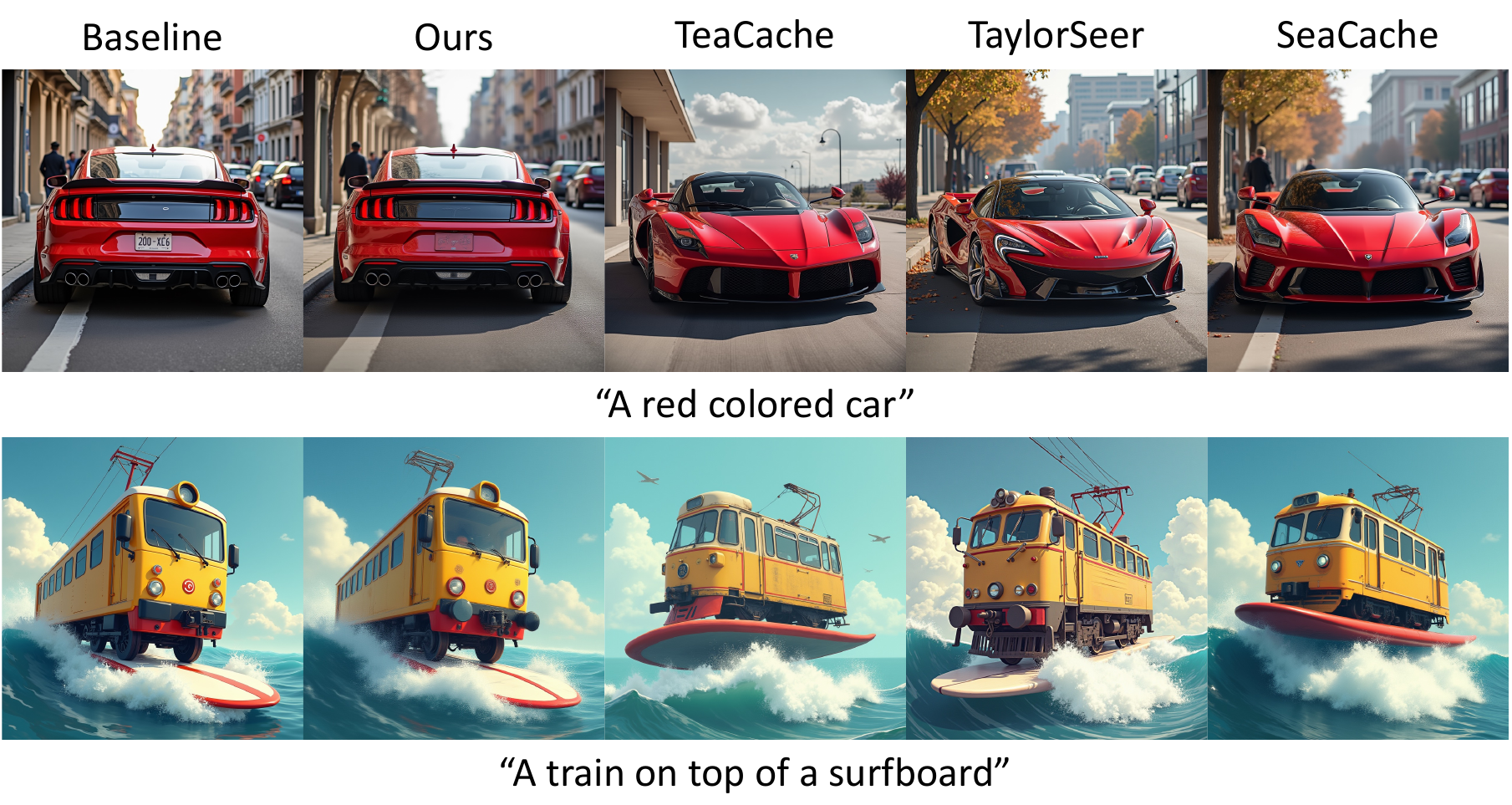}
\caption{Supplementary qualitative comparison on FLUX.1-dev. Each row compares
the uncached baseline with EpaCache (Ours), TeaCache, TaylorSeer, and SeaCache
using the same configuration.}
\label{fig:supp_flux}
\end{figure*}

To complement the quantitative evaluation, Figure~\ref{fig:supp_flux}
presents representative FLUX.1-dev generations produced by EpaCache and
existing caching methods. Compared with the uncached baseline, EpaCache more
faithfully preserves the reference composition and object geometry: it retains
the rear-view vehicle layout in the first example and the train--surfboard
spatial relationship, silhouette, and viewpoint in the second. These examples
illustrate that explicitly accounting for error propagation helps reduce
visible structural drift under cache reuse.

\end{document}